%% file: 0_main.tex
\documentclass[10pt,twocolumn,letterpaper]{article}

\usepackage{cvpr}
\usepackage{times}
\usepackage{epsfig}
\usepackage{graphicx}
\usepackage{amsmath}
\usepackage{amssymb}
\usepackage{subfigure}

\usepackage[pagebackref=true,breaklinks=true,letterpaper=true,colorlinks,bookmarks=false]{hyperref}

\cvprfinalcopy 

\def\cvprPaperID{2455} 

\ifcvprfinal\fi

\begin{document}
\title{SimulCap : Single-View Human Performance Capture with Cloth Simulation}

\author{
	Tao Yu\textsuperscript{1,2},
	Zerong Zheng\textsuperscript{2},
	Yuan Zhong\textsuperscript{2},
	Jianhui Zhao\textsuperscript{1},
	Qionghai Dai\textsuperscript{2},
	Gerard Pons-Moll\textsuperscript{3},
	Yebin Liu\textsuperscript{2}
	\\
	\textsuperscript{1}Beihang University, Beijing, China
	\quad
	\textsuperscript{2}Tsinghua University, Beijing, China
	\\
	\textsuperscript{3}Max-Planck-Institute for Informatics, Saarland Informatics Campus
}

\maketitle

\begin{abstract}
This paper proposes a new method for live free-viewpoint human performance capture with dynamic details (e.g., cloth wrinkles) using a single RGBD camera. Our main contributions are: (i) a multi-layer representation of garments and body, and (ii) a physics-based performance capture procedure. We first digitize the performer using multi-layer surface representation, which includes the undressed body surface and separate clothing meshes. For performance capture, we perform skeleton tracking, cloth simulation, and iterative depth fitting sequentially for the incoming frame. By incorporating cloth simulation into the performance capture pipeline, we can simulate plausible cloth dynamics and cloth-body interactions even in the occluded regions, which was not possible in previous capture methods. Moreover, by formulating depth fitting as a physical process, our system produces cloth tracking results consistent with the depth observation while still maintaining physical constraints. Results and evaluations show the effectiveness of our method. Our method also enables new types of applications such as cloth retargeting, free-viewpoint video rendering and animations.
\end{abstract}

\input{1_Introduction.tex}
\input{2_RelatedWork.tex}
\input{3_AvatarDigitalization.tex}
\input{4_SimulPerfCap.tex}
\input{5_Results.tex}
\input{6_Conclusion.tex}

{\small
\bibliographystyle{ieee}
\bibliography{0_main.bib}
}

\end{document}

%% file: 1_Introduction.tex
\section{Introduction}
Real-time human performance capture using a low budget and an easy setup (e.g., a single depth camera) is a challenging but important task. Fulfilling this goal may enable many applications such as augmented reality, holography telepresence, virtual dressing, etc. 
Recent advances in single-RGBD 4D reconstruction ~\cite{newcombe2015dynamic, tao2018DoubleFusion, habermann2019TOG} have enabled \emph{capture} of geometry, motion, texture and even surface albedo ~\cite{guo2017real} of human performances. However, results are still far from realistic.

The lacking of realism in existing capture methods (in great part) is due to the following limitations.  
First of all, they all use single piece of geometry for reconstruction (the observed human skin and dressed cloth are connected) so they cannot track and even describe cloth-body interactions, such as layering and sliding. Moreover, the reconstructed results are not editable and animatable, which is very important for many applications like virtual dressing. 
Second, clothing can not be described by the typically used kinematic chains or sparsely sampled non-rigid deformation node graphs ~\cite{newcombe2015dynamic, guo2017real, tao2018DoubleFusion}, which leads to degraded capture accuracy and over-smoothed results.
Third, they can not capture the dynamic deformations of the \emph{occluded} part of the clothed person using single-view setups. 

Cloth simulation methods are at the other side of the spectrum; those can generate plausible cloth dynamics on top of a moving body ~\cite{terzopoulos1987elastically,selle2009robust,jiang2017anisotropic}. 
The problem here is adjusting the parameters to achieve a desired realistic animation. Furthermore, complex soft-tissue motions, and cloth-body interactions are extremely difficult to formulate using physics--even when allowing for long processing times.  

\begin{figure}
    \centering
    \includegraphics[width=1.0\linewidth]{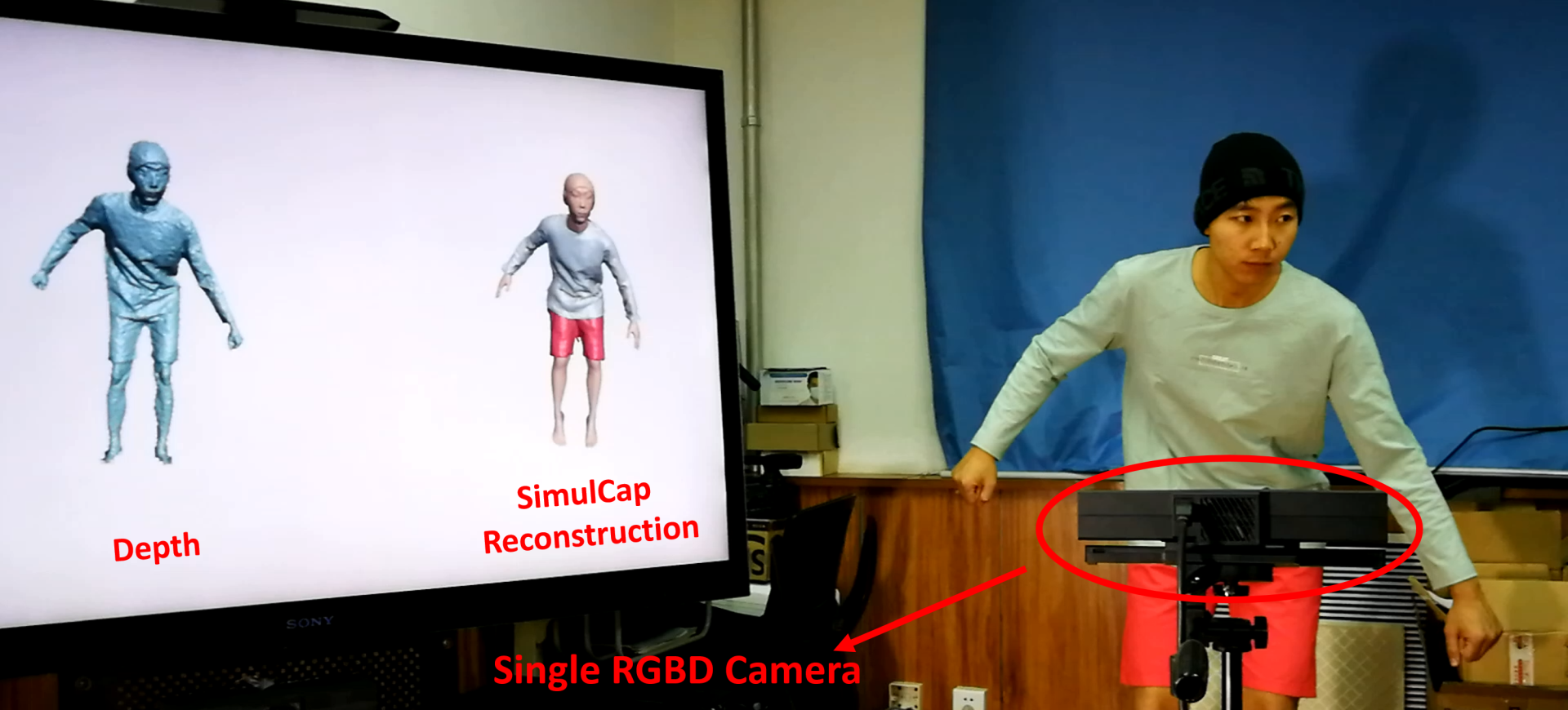}
    \caption{System setup and live reconstruction results. }
    \label{fig:teaser}
\end{figure}


In this work, we introduce SimulCap, a single view RGBD live capture method that combines a layered representation of the human body and clothing with physics based simulation; SimulCap captures cloth motion separately from the body, generates cloth dynamics which satisfy physical constraints, recovers the motion of the occluded parts, and produces a simulation that matches the observed data. 
SimulCap uses a multi-layer surface representation automatically extracted from the data; this is needed for cloth simulation and to model cloth layering. Our observation is that cloth deformation is caused mostly due to the skeletal motion of the underlying body, which is easier to capture.  
By simulating the cloth on top of the captured human body, we achieve two goals: (i) we obtain a good initialization for data fitting of the visible part, and (ii) we can predict the occluded cloth part more accurately than the commonly used surface skinning ~\cite{Baran2007skin} and non-rigid warping ~\cite{li2009robust}.
To capture detail beyond simulation, the observed cloth details, such as wrinkles, can be reconstructed by formulating data fitting as a physical process, which is not only much more efficient, but also preserves physical constraints in the cloth simulation step. 

For single-view live capture of human performances, our method can reconstruct \emph{realistic} results at the visible region and \emph{plausible} results at the invisible area. SimulCap consists of two stages: multi-layer avatar digitization and physics-based performance capture as shown in Fig.\ref{fig:overview}.
We first automatically digitize the subject into a multi-layer avatar, which includes separate surfaces of the undressed body and each of the garments--the subject only needs to turn around one time in front of the camera at the beginning.
During the performance capture step, we track both skeleton motion of the undressed body and the detailed non-rigid deformation of the cloth sequentially. By combining cloth simulation with iterative depth fitting, we can achieve realistic performance capture results. We also demonstrate the flexible ability of our system in cloth retargeting application. 
In summary, SimulCap combines the benefits of capture and simulation, and it constitutes the first live capture method capable of tracking human body motion and clothing separately, while incorporating physical constraints.

%% file: 2_RelatedWork.tex
\section{Related Work}

\paragraph{Human Performance Capture.} 
A large body of works require a \emph{pre-scanned template} to model the body and the clothing using a \emph{single surface} (~\cite{Aguiar08performancecapture,vlasic2008articulated,gall2009motion,liu2011markerless,taylor2012vitruvian,Pons-Moll:IJCV:2015,ye2012performance,ye2014real,li2009robust,guo2015robust,MonoPerfCap_SIGGRAPH2018}). Aguiar \emph{et al.}~\cite{Aguiar08performancecapture}, Vlasic \emph{et al.}~\cite{vlasic2008articulated}, Gall \emph{et al.}~\cite{gall2009motion} and Liu \emph{et al.}~\cite{liu2011markerless} demonstrated high quality performance capture from multi-view video input. Ye \emph{et al.}~\cite{ye2014real} embedded an articulated deformation model into a Gaussian Mixture Model for skeleton tracking. Li \emph{et al.}~\cite{li2009robust} embedded a deformation graph~\cite{Sumner2007embededed} and tracked non-rigid surface motion imposing a local as-rigid-as-possible constraint. Xu \emph{et al.}~\cite{MonoPerfCap_SIGGRAPH2018} combined skeleton motion, non-rigid deformation and 3D pose detection for performance capture from monocular RGB video. Habermann \emph{et al.}~\cite{habermann2019TOG} further demonstrates real-time capture from monocular RGB video. Although these methods can achieve very good performance, they require pre-acquisition of a template; furthermore, clothing and body are represented as a single connected surface, which limits their ability to capture detailed clothing deformations and cloth-body interactions.

Parametric body models can be used to bypass the requirement of a pre-scanned templates. Chen \emph{et al.}~\cite{chen2016realtime} adopted SCAPE~\cite{Anguelov2005scape} to track skeleton motion from a single depth camera. Bogo \emph{et al.}~\cite{BogoBL015Detailed} extended SCAPE to capture detailed body shape (without clothing) with appearance during skeleton tracking. Capturing human shape and pose from an RGB image is much more challenging and ill-posed due to the lack of depth cues. Bogo \emph{et al.}~\cite{Bogo2016keepsmpl} constraints the problem using SMPL~\cite{Loper2015SMPL} to fit predicted 2D joint locations, while ~\cite{hmrKanazawa17,omran2018neural,pavlakos2018humanshape,tung2017self} estimated shape and pose by integrating SMPL as a layer in a CNN-based framework. 
Other works focus on estimating body shape \emph{under} clothing ~\cite{zhang2017detailed,yang2016estimation,wuhrer2014estimation}.
These works are restricted to the shape space of the body model, which can not represent personalized detail, clothing and hair.
Recently, Alldieck~\emph{et al.}~\cite{alldieck2018video,alldieck2018detailed,alldieck2019learning} reconstruct clothing and hair, represented as displacements on top of SMPL, from an RGB video of a person. The motions are restricted to rotating around the camera.
Using an RGBD sensor, DoubleFusion ~\cite{tao2018DoubleFusion} achieved highly robust and accurate capture for a variety of motions by combining SMPL with a voxel representation to represent clothing. None of these methods can separate each of the garments from the body, nor predict cloth deformations for the occluded parts on the RGB/depth video. 

Another branch of work focuses on reconstructing the geometry and motion of non-rigid scenes simultaneously. Collet \emph{et al.}~\cite{Collet2015FVV} reconstruct high quality 4D sequences using multi-view setup and controlled lighting. Fusion4D ~\cite{dou2016fusion4d} and Motion2Fusion ~\cite{dou2017motion2fusion} set up a rig with several RGBD cameras to capture dynamic scenes with challenging motions in real-time. DynamicFusion like approaches ~\cite{newcombe2015dynamic, innmann2016volume, guo2017real, slavcheva2017cvpr, tao2017BodyFusion,slavcheva2018cvpr,  chao2018ArticulatedFusion} reconstructed geometry and non-rigid motion simultaneously form a single-view and in real-time. 
The aforementioned methods either require multi-camera setups or can not capture the occluded regions.

\begin{figure*}
	\begin{center}
		\includegraphics[width=0.95\linewidth]{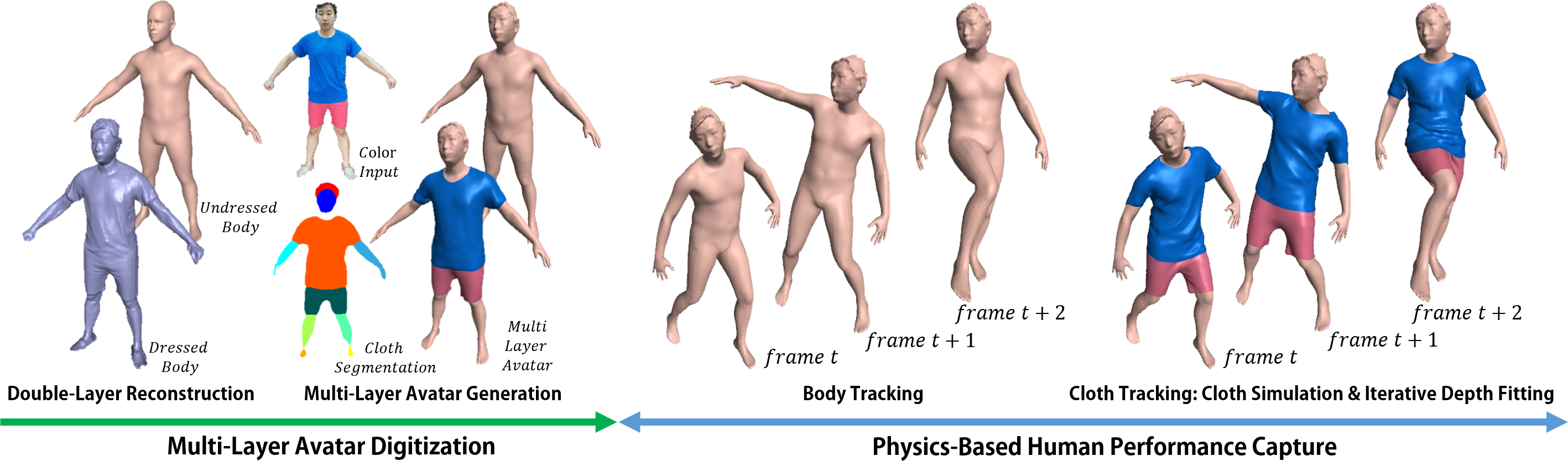}
	\end{center}
	\caption{The pipeline of our system. The first step is to get a multi-layer avatar using \emph{Double-Layer Surface Reconstruction} and \emph{Multi-Layer Avatar Generation}. Then we capture human performances by performing \emph{Body Tracking} and \emph{Cloth Tracking}, which include cloth simulation and iterative depth fitting, sequentially for each new RGBD frame. The final capture process is: turn around in front of a camera, wait several seconds for multi-layer avatar digitization and then start physics-based human performance capture in real-time. }
	\label{fig:overview}
\end{figure*}

\paragraph{Cloth Simulation and Capture.} Works in this category model or capture clothing more explicitly.
Simulation of clothing has been investigated for more than 30 years. Some works focus on super realistic cloth simulation results using millions of triangles ~\cite{terzopoulos1987elastically,selle2009robust,jiang2017anisotropic}, while the others concentrate on improving the fidelity for real-time simulation \cite{provot1995deformation,muller2007PBD,muller2008hpbd,goldenthal2007efficient,gillette2015real,kavan2011physics,kim2013near,wang2010example,de2010stable,guan2012drape,xu2014sensitivity,selle2009robust}. Physics-based mass-spring models \cite{provot1995deformation} or position based dynamics ~\cite{muller2007PBD,muller2008hpbd} are commonly used for simulation of cloth. 
Realism, controlability and speed remain open problems for simulation methods.
Example-based methods ~\cite{kim2013near,wang2010example,de2010stable,xu2014sensitivity,guan2012drape} learn from offline animations to achieve real time performance, but generalization to novel motions, shapes and fabrics is challenging.

Static cloth capture has been demonstrated, to some degree, from single images~\cite{zhou2013garment,danvevrek2017deepgarment} or RGBD~\cite{chen2015garment}.
Reconstruction typically requires manual intervention, or learning for a specific set of garments. 
Dynamic cloth reconstruction~\cite{Bradley:2008,Popa:2009} typically requires multi-view studios, and is restricted to capturing a single garment, and not the person wearing it.
Both simulation and capture can be leveraged by estimating the physics parameters of cloth~\cite{rosenhahn2007system,stoll2010video} or soft-tissue~\cite{Meekyoung:siggraph} from multi-view or 4D captured results, with the goal of driving simulation for new motions. Instead, we use simulation \emph{during} capture to reconstruct occluded deformations from a \emph{single} camera.
Data-driven models are alternative to simulation; they learn how the clothing deforms on top of the body~\cite{neophytou2014layered,yang2018analyzing,Lahner_2018_ECCV}. Although this is a promising direction, the models can not separate the garments from the body~\cite{neophytou2014layered,yang2018analyzing}, or require garment specific learning~\cite{Lahner_2018_ECCV}.

The most relevant here is the work of Pons-Moll~\emph{et al.}~\cite{Pons-Moll:Siggraph2017}(ClothCap). 
Similar to us, they jointly estimate body shape, pose and cloth deformation by using separate meshes for garments and body. However, they only demonstrate results using 4D scans as input, which do not suffer from occlusion. We address the more challenging monocular RGBD scenario, and show how physics-based simulation can help to capture the non-visible parts.

%% file: 3_AvatarDigitalization.tex
\section{Multi-Layer Avatar Digitization}
\label{sec:multi_layer_avatar_digitization}
Multi-layer avatar digitization consists of two steps: double-layer surface reconstruction and multi-layer digitization. Double-layer surface comprises the surface of the dressed body and the undressed body as shown in Fig.~\ref{fig:overview}. 

We obtain a double-layer surface using DoubleFusion~\cite{tao2018DoubleFusion}, which is a single-view, real-time method, which reconstructs the dressed body and undressed body surface at the same time. People only need to turn around once in front of a depth camera. To obtain a complete dressed surface without holes, we perform Poisson surface reconstruction ~\cite{Kazhdan:2006:PSR:1281957.1281965} and remeshing ~\cite{Jakob2015Instant}. This ensures a complete manifold for later segmentation and cloth simulation steps. 

For multi-layer avatar generation, we parse and segment different cloth from the dressed body surface. However, cloth segmentation is a difficult task even for 2D images, so we require the colors of garment to be sufficiently different. By combining a learning based image parsing and volumetric fusion together we can get robust 3D cloth parsing and segmentation results efficiently and automatically. 

To initialize the segmentation step, we use the state-of-the-art learning-based human parsing algorithm ~\cite{LIP2018} to get cloth segmentation at the first frame as shown in Fig.\ref{fig:overview}. Then we estimate cloth colors using K-means and use it to segment the rest input rgb frames. We fuse all the color segmentation results into a parsing volume (A volume has the same resolution and size as the TSDF volume) which is not only robust to noisy 2D segmentation, but also very convenient to segment 3D clothing under the volume representation. Specifically, for each parsing voxel inside the truncated band of the TSDF volume, we first project it onto each input RGB image according to the tracked non-rigid motions, and store the corresponding pixel segmentation labels. 
The value saved in each parsing voxel is an array of label frequencies. Note that we only consider $3$ labels (upper cloth, lower cloth and skin) in order to simplify the segmentation. In addition to label frequencies, we also fuse color into the parsing volume for subsequent segmentation on the surface using MRF. After cloth segmentation, we smooth the noisy boundaries of the segmented cloth pieces and handle the occlusion between multiple clothes by assuming that upper cloth is always outside pants. 

In order to augment the realism of the reconstructed avatar, we enhance the head of the undressed body by deforming it to fit the fused head on the dressed surface.

%% file: 4_SimulPerfCap.tex
\section{Physics-Based Performance Capture}
\label{sec:performance_capture}
The physics-based human performance capture contains $2$ steps: body tracking and cloth tracking. In the first step, we track the motion of the undressed body. In the second step, detailed cloth motion is tracked based on both, cloth simulation, and current depth input. 
\subsection{Body Tracking}
\begin{figure}
    \centering
    \includegraphics[width=1.0\linewidth]{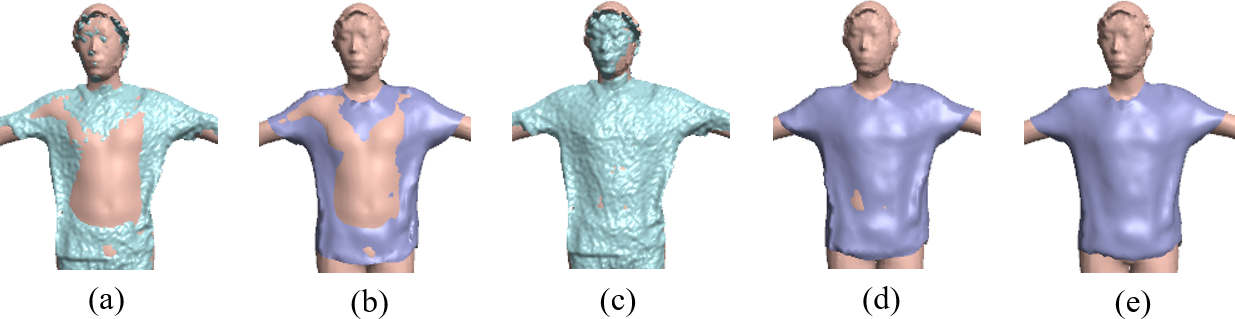}
    \caption{Illustration of interpenetration term. (a,c) Overlay between depth and tracked body without/with using interpenetration term; (b) and (d) are cloth tracking results based on (a) and (c) by direct depth fitting; (e) Cloth tracking results based on (c) using SimulCap. With interpenetration term, we can generate more realistic body and cloth tracking results. }
    \label{fig:illu_interpenetration}
\end{figure}

The challenge of body tracking in our system is that we have to track accurate skeleton motion of the undressed body only given the depth of the dressed body. Moreover, the body-depth interpenetration during skeleton tracking, which is not considered in previous methods, is a very important factor in our system. This is because severe interpenetration will deteriorate the subsequent cloth tracing step as shown in Fig.\ref{fig:illu_interpenetration}.

Iterative closest point algorithm (ICP) is used for skeleton tracking. To eliminate the ambiguity between undressed body and dressed-depth-input, we leverage the reconstructed double layer surface (in Sec.\ref{sec:multi_layer_avatar_digitization}) for constructing the tracking data term as in ~\cite{tao2018DoubleFusion}. 
Moreover, we construct another interpenetration term to limit body-depth interpenetration. The energy function of body tracking is: 
\begin{equation} \label{eqn:skel_tracking}
E_{\mathrm{skel}} = 
\lambda_{\mathrm{data}}E_{\mathrm{data}} + 
\lambda_{\mathrm{inter}}E_{\mathrm{inter}} + \lambda_{\mathrm{pri}}E_{\mathrm{pri}},
\end{equation}
where the $E_{\mathrm{data}}$ measures the fitting between the skinned double layer and the input depth point cloud, please refer to ~\cite{tao2018DoubleFusion} for detailed formulation;
$E_{\mathrm{inter}}$ measures body-depth interpenetration; 
$E_{\mathrm{pri}}$ is human pose prior in ~\cite{Bogo2016keepsmpl} for penalizing unnatural poses. 

The interpenetration term is defined as:
\begin{equation}
\label{eqn:skel_inter}
E_{\mathrm{inter}} = 
\sum_{(v_b, u_c)\in\mathcal{Q}}
{|\mathbf{v_b} - (\mathbf{u_c} - \mathbf{n}_{u_c}\sigma)|^2},
\end{equation}
where $\mathcal{Q}$ is the correspondence set of all the interpenetrated body vertices $\mathbf{v_b}$ and its nearest cloth depth point $\mathbf{u_c}$; $\mathbf{n}_{u_c}$ is the normal of $\mathbf{u_c}$; $-\mathbf{n}_{u_c}\sigma$ represent a shift along the inverse normal direction to make sure the target position of $\mathbf{v_b}$ is behind the depth observation. By incorporating interpenetration term, we can get better body tracking and cloth tracking results as shown in Fig.~\ref{fig:illu_interpenetration}. 

\begin{figure*}
	\begin{center}
		\includegraphics[width=1.0\linewidth]{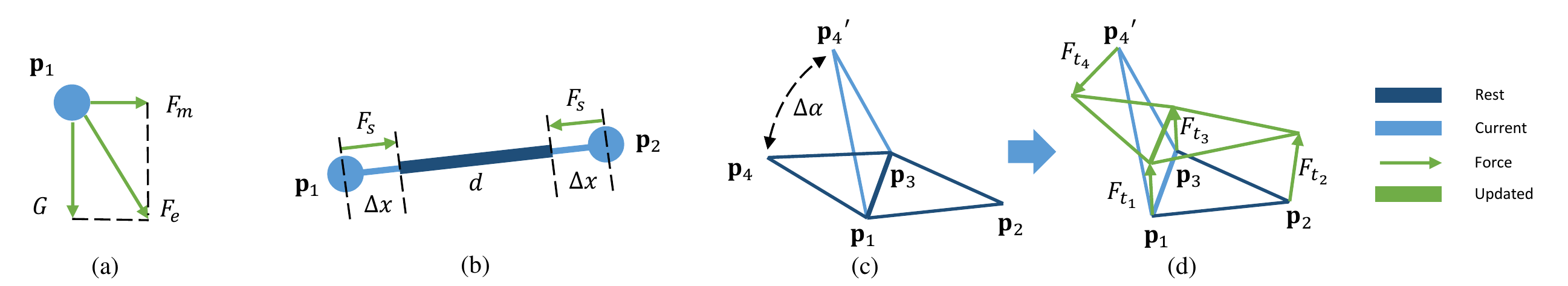}
	\end{center}
	\caption{Simulation forces in our system. (a) Resultant external force $F_e$ on vertex $\mathbf{p}_1$, which include gravity $G$ and other external force $F_m$; (b) Stretching force $F_s$ between vertex $\mathbf{p}_1$ and $\mathbf{p}_2$, where $d$ is the rest length of the stretching spring and $\Delta{x}$ is half of the length difference between rest length and current length; (c) The rest status and delta angle $\Delta{\alpha}$ of current status of torsion spring. (d) Torsion force on the $4$ vertices ($\mathbf{p}_1$, $\mathbf{p}_2$, $\mathbf{p}_3$ and $\mathbf{p}_4$) of the two connected triangles and the updated position.}
	\label{fig:forces}
\end{figure*}


\subsection{Cloth Tracking}
\begin{figure}
	\begin{center}
		\includegraphics[width=1.0\linewidth]{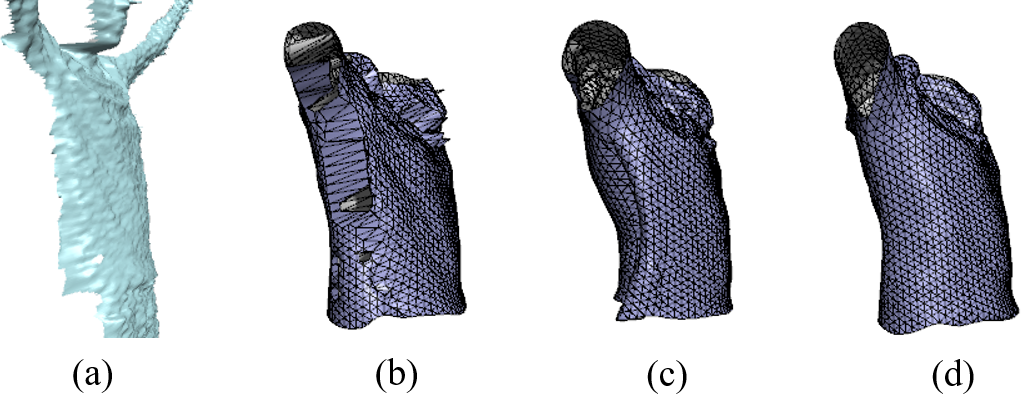}
	\end{center}
	\caption{Illustration of iterative depth fitting. From left to right: depth input (a), result of direct depth fitting (b), iterative depth fitting without (c) and with (d) smooth blending.}
	\label{fig:iterative_depth_fitting}
\end{figure}

Given the undressed body with its motion, we can simulate plausible cloth dynamics for both visible and invisible regions. The simulated cloth provides very good initial status which facilitate fitting the cloth to the depth input. However, the depth fitting process remains non-trivial. 
It is difficult for previous methods like Laplacian Deformation and non-rigid registration to achieve wrinkle-level detailed deformations under a real-time budget and locally as-rigid-as-possible constraints. 
Moreover, stretching the simulated cloth to the depth input directly may generate many artifacts: First, since we can only get partial observation under single-view setup, there must be a gap between the simulated region and the depth fitting region on the cloth, which will leads to spatial discontinuity on the final reconstructed cloth mesh as shown in Fig.~\ref{fig:iterative_depth_fitting}(b)(depth boundary region); 
Second, the direct depth fitting method may even break the consistency of the internal physical constraints and generate non-physical fitting results as shown in Fig.~\ref{fig:iterative_depth_fitting}(b)(chest region), which will also lead to unexpected simulation results in the next frame. 
So we propose a method that performs depth fitting iteratively as a physical process, which can not only achieve efficient\&realistic depth fitting, but also maintain the physical constraints in the simulation step. 

\subsubsection{Cloth Simulation}
For cloth simulation, although a lot of advanced cloth simulation methods have been proposed in recent years, we extend classical Force-Based Mass-Spring method ~\cite{provot1995deformation} due to its efficiency and simplicity. This method models cloth as a mass-spring system in which each vertex has a mass and all the vertices are connected by springs. Force-based methods calculate the resultant force for each vertex (mass) explicitly. The simulation steps can be concluded as: 

\begin{itemize}
\item Initialization: Calculate the initial status of the mass-spring system (assign mass for each vertex and calculate rest status for all the springs);
\item Simulation: For each vertex in each time step:
    
    1) Resultant Force Calculation. Calculate the sum of all the internal and external forces for each vertex, and then calculate vertex acceleration according to the fundamental law of dynamics. The internal forces are generated by different types of springs (internal constraints) while the external forces include omnipresent loads (e.g., gravity) and other specified external forces. 
    
    2) Vertex Position Update. Using Explicit Euler Integration to update the position of each vertex according to the resultant vertex acceleration.  
    
    3) Collision Handling. Detect and handle different types of collision for cloth and body. 
\end{itemize}

We regard all the edges on the triangle mesh as stretching springs and they provide in-plane constraints. And to constrain the bending of the triangle mesh, we add an additional torsion spring on the common edges of two connected triangles. By adjusting the stiffness of different springs, we can approximate different types of cloth materials in the real world. We illustrate all the forces of our system below. 


The resultant external force can be calculated as the summation of all the external forces, $F_e = G + F_m$, as shown in Fig~\ref{fig:forces}(a). Note that we use the inverse normal direction of the floor (which we detected at the first frame) as the direction of gravity in our system.


We suppose all vertex have the same mass. The stretching force $F_s$ on the two vertices of stretching spring are equal and can be calculated as: 
$F_s = \mathbf{k} * \Delta{x}$, 
where $\mathbf{k}$ is the stiffness of the stretching spring and $\Delta{x}$ is the half of length difference between current length and rest length of the stretching spring (Fig.~\ref{fig:forces}(b)).

The moment of torsion spring is defined as $\mathbf{M} = \mathbf{w}*\Delta{\alpha}$, 
where $\mathbf{w}$ is the torsion coefficient and $\Delta{\alpha}$ is the angle difference between current angle and rest angle of the torsion spring (Fig.~\ref{fig:forces}(c)). The angle of the torsion spring was calculated by the angle between the normal vectors of the two connected triangles. We calculate torsion force $F_{t_i}$ on each vertex (Fig.~\ref{fig:forces}(d)) according to the bending constraints projection method in ~\cite{muller2007PBD}. 


We suppose all the stretching spring has the same stretching coefficient $\mathbf{k}$ while all the torsion spring has the same stiffness coefficient $\mathbf{w}$ for simplicity. We implement the simulation method on GPU. Kernel merge and warp shuffle techniques are used to further improve performance. The time step is set to $0.00033s$ to avoid the overshooting and unstable problems of explicit Euler integration as illustrated in ~\cite{provot1995deformation}. We perform explicit Euler integration 100 times between two rendering frames and linearly interpolate the position of each inner body vertex for body-cloth collision handling. We use static collision handling scheme for body-cloth collision while use continuous collision handling scheme for cloth-cloth collision as in ~\cite{provot97collision}. 

\subsubsection{Iterative Depth Fitting}
Although the simulated cloth is plausible and consistent in both visible and invisible regions, it by no means exactly the same with the real-world cloth because real physical world always have much more factors that is hard to simulate, for example, the weave and structure of the cloth or even soft tissue motions of the undressed body etc. However, for performance capture systems, the goal is to capture \emph{real world performances} efficiently, so the captured results should fit the \emph{real observations}. Therefore, we need to fit the visible area of the simulated cloth to current depth input for generating more realistic cloth tracking results.

\begin{figure}
	\begin{center}
		\includegraphics[width=1.0\linewidth]{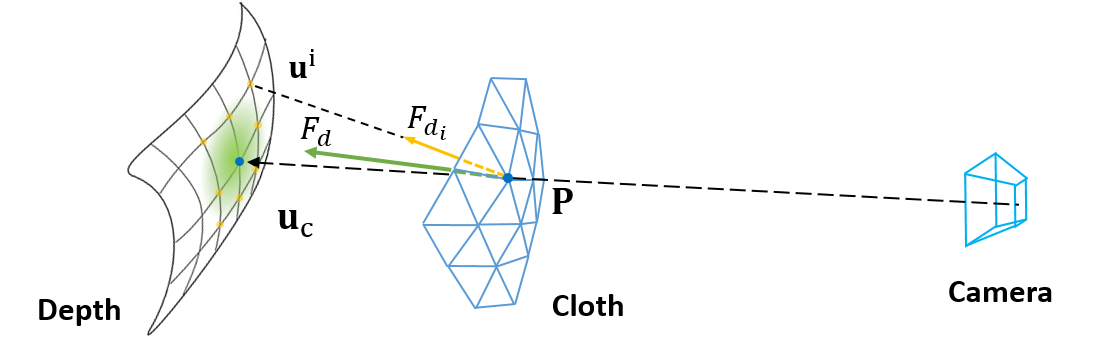}
	\end{center}
	\caption{Illustration of depth fitting force $F_d$.}
	\label{fig:depth_fitting_force}
\end{figure}

\begin{figure}
	\begin{center}
		\includegraphics[width=1.0\linewidth]{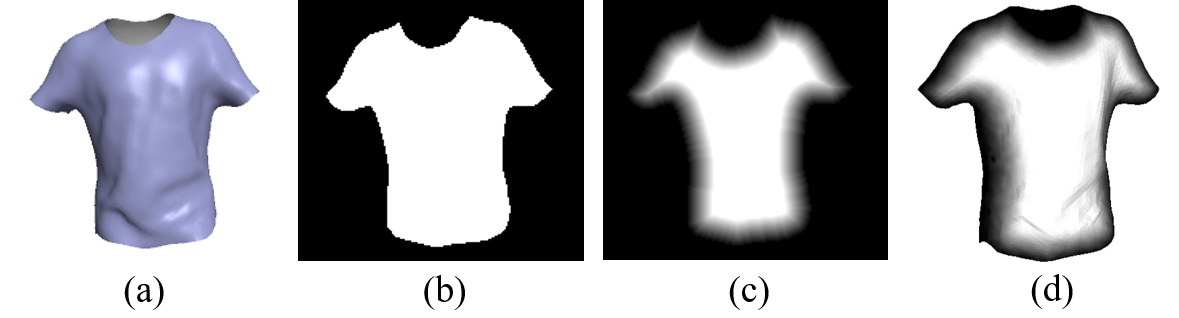}
	\end{center}
	\caption{Illustration of smooth blending mask generation. The first step is to render the mask image (b) and calculate the visibility of the cloth mesh based on the simulated cloth (a). Then calculate the distance transform of (b) and calculate the 2D smooth blending mask (c). Finally, each visible cloth vertex acquire their depth fitting weight by projecting to the 2D blending mask. The final color coded cloth mesh is in (d) with depth fitting weight from $0$ (black) to $1$ (white), note the smooth transition of depth fitting weight around the boundary of the visible area. }
	\label{fig:blending_mask}
\end{figure}

Inspired by the physics-based simulation algorithms, we formulate the depth fitting process as a physical process, in which the input depth point has attraction to the cloth. Thus, a new force should be defined and has positive correlation to the distance between cloth and depth.
We name the new force as depth fitting force which only affects cloth vertices. The force is defined as: 
\begin{equation}
    F_d = 
    \psi(\mathbf{p})\cdot
    \sum_{u^i\in\mathcal{N}(u_c)}
    {\eta \cdot \tau(\mathbf{u_c}) \cdot
    e^{\gamma|\mathbf{u}^i - \mathbf{P}|}\cdot \frac{\mathbf{u}^i - \mathbf{P}}{| \mathbf{u}^i - \mathbf{P}|}},
\end{equation}
where $\mathbf{P}$ is a visible cloth vertex; $\mathbf{u_c}$ is the projective depth point of $\mathbf{P}$; $\mathbf{u^i}$ is the $i$th neighbor vertex in the 1-ring neighbor set $\mathcal{N}(u_c)$ of $\mathbf{u_c}$; $\tau(\mathbf{u^i})$ is a 2D gaussian kernel defined on $\mathbf{u_c}$ for blending the fitting forces $F_{d_i}$ generated from all the neighbor points as shown in Fig.~\ref{fig:depth_fitting_force}. $\psi(\mathbf{P})$ is the smooth blending weight for depth fitting which we described in Fig.\ref{fig:blending_mask};
$\gamma$ and $\eta$ are scaling factors to keep the force in a valid range and we set it to $240$ and $0.34$ respectively. 


Iterative depth fitting is then achieved by performing cloth simulation again, in which we consider the depth fitting force as external force $F_m$ in Fig.~\ref{fig:forces}(a) and the undressed body be static during the simulation. The incorporation of physical constraints into the depth fitting process can not only keep the consistency of internal physical constraints in the simulation step, but also act as a physical filter to eliminate non-physical observations (e.g., large noise) on the input depth. Note that the displacement of the depth fitting step should not produce additional velocities to the cloth in the next cloth simulation step. 

Even we ``simulate'' the depth fitting process, we may still have spatial discontinuities around the depth boundaries as shown in Fig.~\ref{fig:iterative_depth_fitting}(c). The reasons are two folds: On one hand, the simulated cloth cannot perfect align with the depth boundaries due to the non-perfect body tracking and cloth simulation results; On the other hand, the depth boundaries always contains much more noise than the central areas which makes situation even worse. To get a smooth transition between simulated region and depth fitting region around depth boundaries, we first generate a smooth blending mask using the 2D silhouette of the simulated cloth. Then scaling the depth fitting force for each cloth vertex according to the value of the smooth blending mask. The more a vertex is close to the boundary, the less it will be forced to fit depth observation. We illustrate the generation of the smooth blending mask in Fig.\ref{fig:blending_mask}. The result of smooth blending is shown in Fig.~\ref{fig:iterative_depth_fitting}(d). We iterate $100$ times for iterative depth fitting as in the cloth simulation step. 

\begin{figure*}
	\begin{center}
		\includegraphics[width=1.0\linewidth]{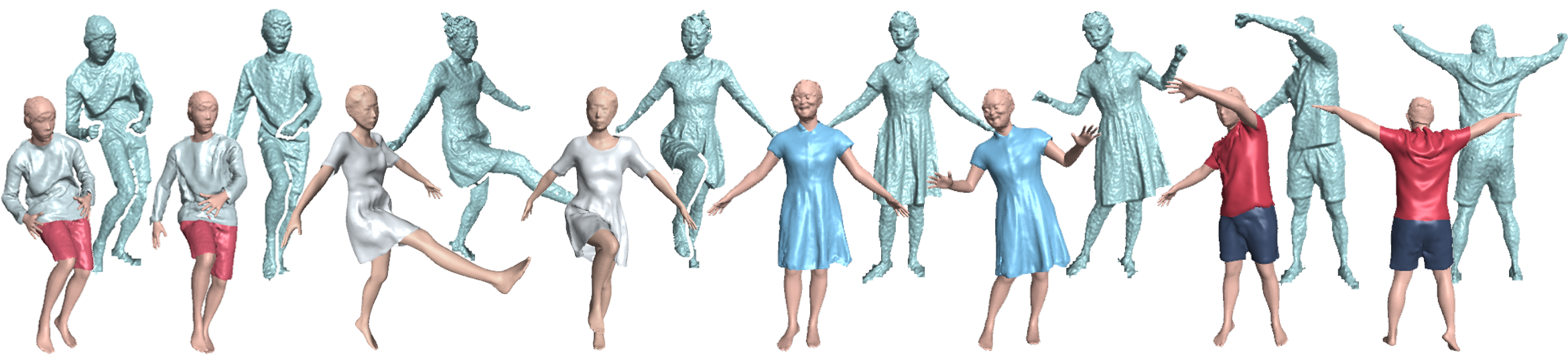}
	\end{center}
	\caption{Reconstruction results of our system.}
	\label{fig:results}
\end{figure*}

Note that we have to perform cloth simulation first for getting a good approximation of current depth input and then perform iterative depth fitting. Other than that, the driving force of the moving body and the depth fitting force may conflict with each other and generate unnatural results. This is the reason why we have to split cloth simulation with iterative depth fitting. 


%% file: 5_Results.tex
\section{Results}
We demonstrate our results in Fig.~\ref{fig:results}. Note the faithful cloth dynamics that we reconstructed. 

SimulCap is implemented on one NVIDIA TITAN Xp GPU. An efficient Gauss-Newton solver was implemented for body motion optimization. We perform cloth simulation in parallel and use kernel merge techniques to further improve the performance. The double-layer surface reconstruction takes $4-6s$ for different self-turning around motions. The multi-layer avatar generation step takes $10s$, with post-geometry-processing $7.9s$( poisson reconstruction $5.5s$ and remeshing $2.4s$), cloth segmentation $1.5s$ and body enhancement $0.5s$. Note that we perform parsing fusion along with the TSDF fusion step in double-layer surface reconstruction to improve efficiency. And the physics-based performance capture step takes $56ms$, with body tracking $17ms$, cloth simulation $14ms$, iterative depth fitting $20ms$ and all the rest steps takes $5ms$. 

For poisson reconstruction, we set the depth of octree as $8$ and the final edge length of each triangle after remeshing is about $2cm$. In body tracking, we perform $6$ ICP iterations, in which $\lambda_{data} = 1.0$, $\lambda_{prior} = 0.01$ and we set $\lambda_{inter} = 1.0$ for the first $3$ iterations and $\lambda_{inter} = 10.0$ for the rest iterations. For cloth simulation, we specify the stretching $\mathbf{k}$ and bending $\mathbf{w}$ parameters for each cloth at the beginning; The velocity damping coefficient in the simulation solver is set to $0.1$ for more stable energy dissipation; The mass of each vertex is set to $0.001$. We classify cloth material into $3$ types: soft ($\mathbf{k} = 300$, $\mathbf{w}=0.01$), middle ($\mathbf{k} = 800$, $\mathbf{w}=0.05$) and hard($\mathbf{k} = 1300$, $\mathbf{w}=0.1$). For each sequence, we choose a type of material parameters according to the real cloth material. 

\subsection{Evaluation}
In this section, we first evaluate our method by comparing with state-of-the-art methods. Then we evaluate the reconstruction in the invisible region in detail, which is a very important improvement of our method. Finally, we evaluate the proposed iterative depth fitting method quantitatively. 

\begin{figure}
	\begin{center}
		\includegraphics[width=1.0\linewidth]{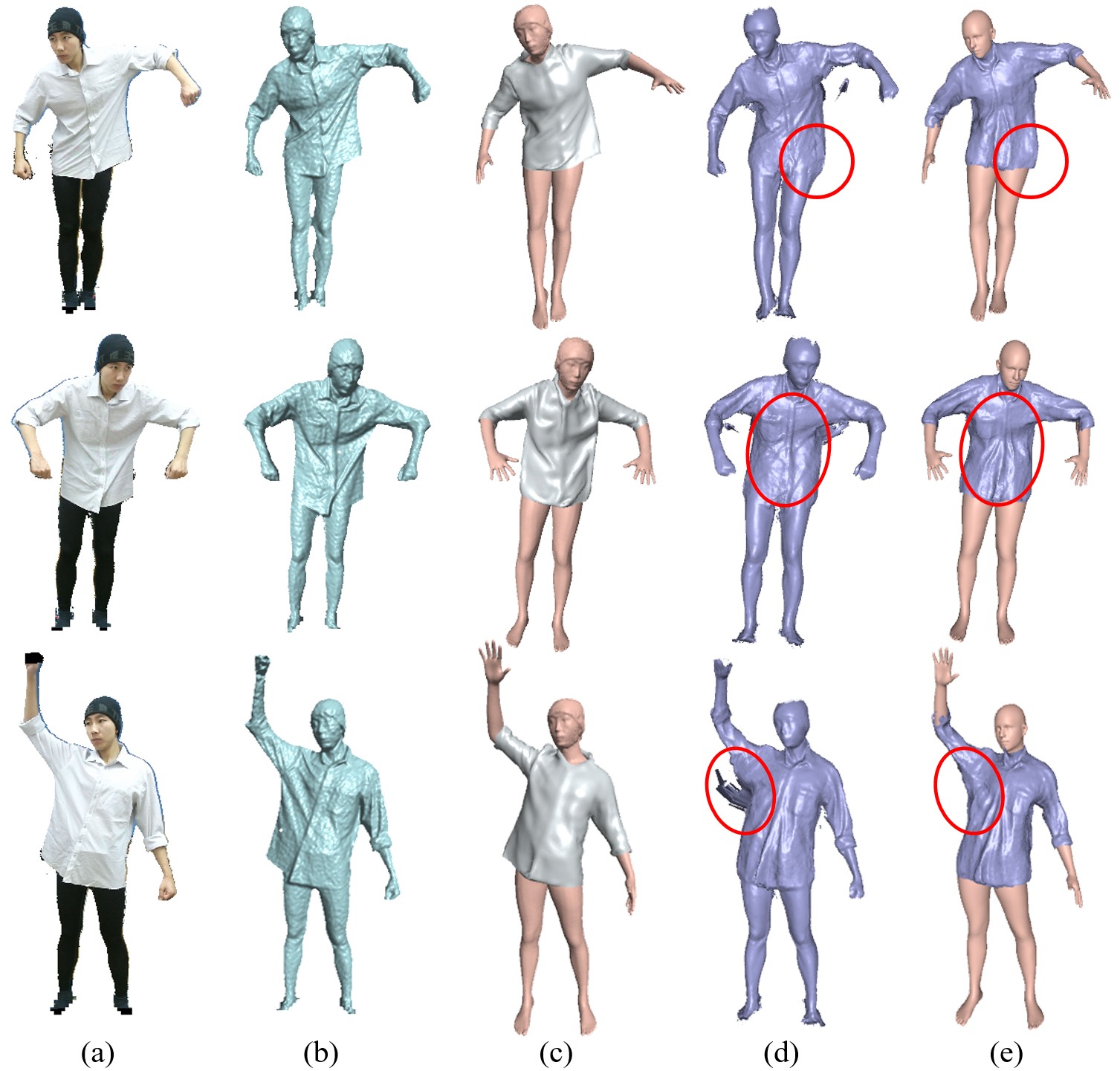}
	\end{center}
	\caption{Comparison with DoubleFusion and the multi-layer baseline method. (a) reference color image (not used during performance capture); (b) depth input; (c) our results; (d) results of DoubleFusion; (e) results of the multi-layer baseline method. }
	\vspace{-10pt}
	\label{fig:comp_dblfu}
\end{figure}

Since we are the first single-view method for live capture of human performances based on multi-layer surface representation, we compare with DoubleFusion, which is the state-of-the-art for single-view real-time human performance capture based on double-layer representation, to demonstrate the effectiveness of our method. Moreover, we also implement a baseline method for multi-layer performance capture, which leverages a multi-layer surface and perform ICP-based non-rigid tracking for each cloth independently. There are $4$ typical improvements of our method:

First, our method achieves much more realistic cloth-body interactions (cloth sliding wrt the body and/or cloth leave the body) as shown in Fig.~\ref{fig:comp_dblfu}(c)(up). DoubleFusion uses a single piece of geometry for representing the outer surface, which means cloth and body are on the same piece of geometry, so they cannot handle naturally separations between cloth and body Fig.~\ref{fig:comp_dblfu}(d)(up). For the multi-layer baseline method, it is still very difficult for ICP-based non-rigid tracking methods to track such challenging interactive motions because of the limited observations (single-view setup) and real-time budget Fig.~\ref{fig:comp_dblfu}(e)(up). 

Second, our method captures much more realistic cloth dynamics compared with the others benefiting from cloth simulation and the  iterative depth fitting scheme Fig.~\ref{fig:comp_dblfu}(c)(middle). The Degree of Freedom (DOF) of non-rigid tracking in DoubleFusion and the multi-layer baseline method is limited due to the real-time budget, so they cannot track detailed cloth dynamics Fig.~\ref{fig:comp_dblfu}(d,e)(middle). Note that in DoubleFusion, it keeps fusing all the depth observation onto the surface, so it may capture cloth details when the subject keep relative still in front of the camera, but it will suffer from delay and fast motion (fast movements will smooth the fused surface details) because it is a temporal fusion process. For the multi-layer baseline method, the geometric details on the cloth, which correspond to the initial static template, do not change over time and, thus, not physically plausible. Moreover, the occluded surface areas is transformed according to skinning/warping alone in DoubleFusion and the multi-layer baseline method, which models mostly articulated deformations. 

Third, a typical artifacts around armpits of DoubleFusion and the multi-layer baseline method Fig.~\ref{fig:comp_dblfu}(d,e)(bottom) can be eliminate by our method. The reason for such artifacts are erroneous surface fusion results, inaccurate skeleton embedding and smooth warping weight around such regions. Our method can generate more plausible results around such regions benefiting from the "divide-and-conquer" scheme as shown in Fig.~\ref{fig:comp_dblfu}(c)(bottom). 

\begin{figure}
	\begin{center}
		\includegraphics[width=1.0\linewidth]{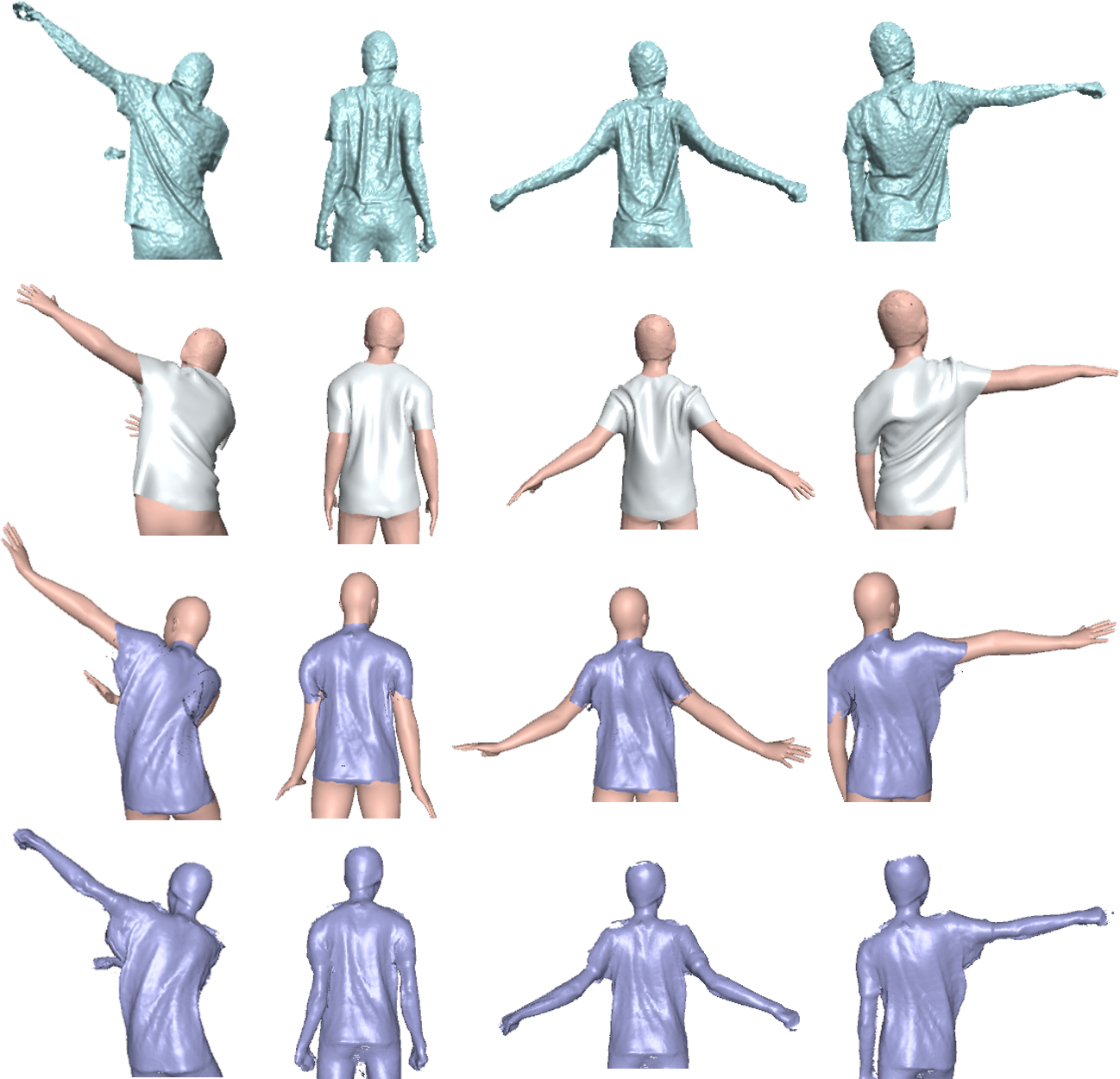}
	\end{center}
	\caption{Evaluation of the reconstruction at the invisible area. The first row are depth observations of the invisible area captured by an additional RGBD camera (not used in any methods). We render the results using a similar view point as the reference depth camera. The rest rows are the results of our method, the multi-layer baseline method and DoubleFusion,  respectively. }
	\label{fig:eval_sim}
\end{figure}

Finally, our method can infer plausible cloth dynamics even in the invisible region Fig.~\ref{fig:eval_sim}(2nd row)(please note the faithful direction and density of the wrinkles we reconstructed), which cannot be achieved by previous methods Fig.~\ref{fig:eval_sim}(3rd and 4th row). 

\begin{figure}
	\begin{center}
		\includegraphics[width=1.0\linewidth]{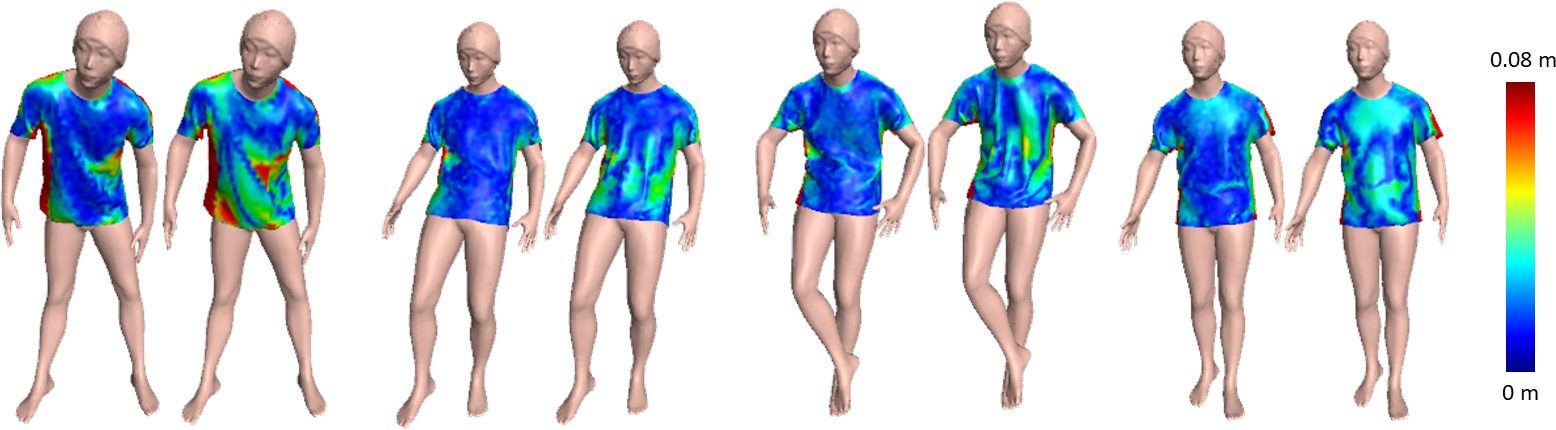}
	\end{center}
	\caption{Evaluation of iterative depth fitting. For each pose, we show our results with (left) and without (right) iterative depth fitting. The depth fitting error is color coded from blue to red.} 
	\label{fig:eval_fitting}
\end{figure}

We evaluate the proposed iterative depth fitting method quantitatively in Fig.~\ref{fig:eval_fitting}. As shown in the figure, with iterative depth fitting, the reconstructed cloth dynamics is more accurate thus much more consistent to the depth input. 

\begin{figure}
    \centering
    \includegraphics[width = 1.0\linewidth]{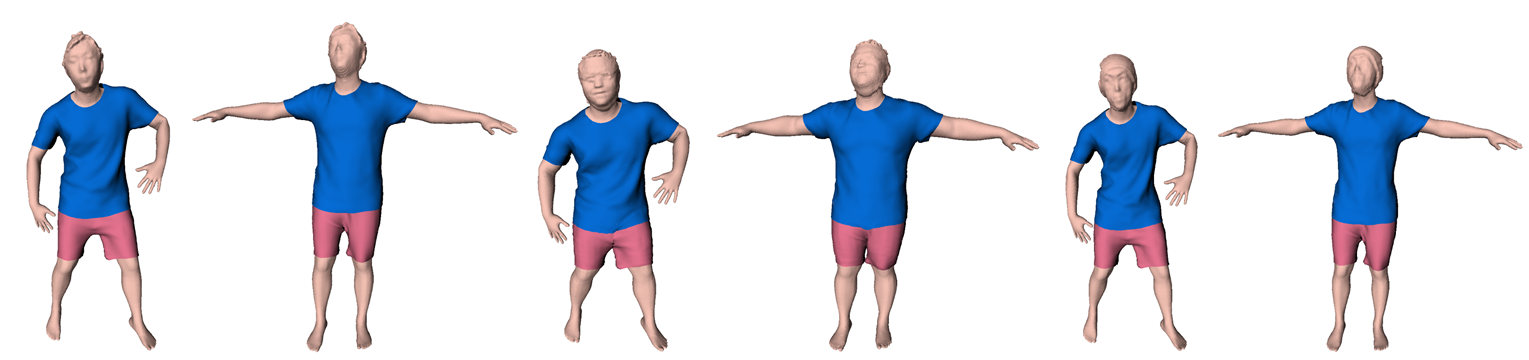}
    \vspace{1pt}
    \caption{Cloth retargeting results from $1$ source subject (left) to $2$ target subjects (middle and right) in different poses.}
    \label{fig:retar}
\end{figure}

\subsection{Applications}
Benefiting from the semantic avatar representation and efficient physics-based performance capture algorithm, we can enable interesting applications like cloth retargeting as shown in Fig.~\ref{fig:retar}. Note that different parts of the multi-layer avatar are well aligned, the undressed body is animatable and the cloth meshes are simulation-ready, so the avatar can be easily incorporated into 3D engines for rendering new free viewpoint sequences. Moreover, much more realistic rendering (with dynamic shading effects) can be achieved given intrinsic texture on the cloth. 


%% file: 6_Conclusion.tex
\section{Conclusion}
We proposed the first method that marries physics-based simulation with performance capture and is capable of tracking people and their clothing using a multi-layer surface. We demonstrated very realistic live capture results using a single-view RGBD camera. 
Higher realism is achieved because our forward model for tracking is closer to how bodies and clothing deform in the real world: Skeletal motion deforms the body, which in turn deforms the clothing layered on top of it. Modelling this process allows us track cloth-body interactions and hallucinate the surface in the occluded regions. We have also demonstrated that this allows to retarget captured clothing to different bodies. In summary, SimulCap demonstrates that modelling the physical process--even using a simple computationally efficient model--allows to capture performances from partial observations.
We believe that this new direction for capture will enable the generation of photorealistic fully-animatable multi-layer avatars for analysis and synthesis, and will open many applications in VR/AR, virtual try-on and tele-presence.

\noindent\textbf{Limitations and Future Work.} 
Although we can reconstruct plausible cloth dynamics even for relatively loose clothing (e.g., skirts), 
the achieved realism in the occluded regions is limited by the quality of the simulator, and tracking of very thick clothing (e.g., sweaters) remains challenging. Incorporating more advanced cloth simulators and take into account the sewing patterns might increase the achieved realism.
Moreover, capturing the natural interactions between hands/arms and cloth requires more accurate physics-based collision models. 
Finally, topology changes, face, hands and soft-tissue can not be faithfully reconstructed using SimulCap, which remains challenging even for multi-view offline methods such as ~\cite{Pons-Moll:Siggraph2017}. Fortunately, capturing clothing and body separately makes it straightforward to integrate new models of faces~\cite{FLAME:SiggraphAsia2017}, hands~\cite{MANO:SIGGRAPHASIA:2017} and soft-tissue~\cite{PonsMoll2015Dyna}. 
Other potential future directions include: Incorporating human soft-tissue models (e.g., \cite{PonsMoll2015Dyna}) to faithfully capture cloth-body interactions, ``learning'' a data-driven clothing deformation model from captured results, and inferring material properties. \\
\noindent \textbf{Acknowledgements} This work is supported by the NSF of China No.61827805, No.61522111, No.61531014, No.61233005; Changjiang Scholars and Innovative Research Team in University, No.IRT\_16R02; Gerard Pons-Moll is funded by the Deutsche Forschungsgemeinschaft (DFG. German Research Foundation)--409792180. 

%% file: 0_main.bbl
\begin{thebibliography}{10}\itemsep=-1pt

\bibitem{Aguiar08performancecapture}
Edilson Aguiar, Carsten Stoll, Christian Theobalt, Naveed Ahmed, Hans peter
  Seidel, and Sebastian Thrun.
\newblock Performance capture from sparse multi-view video, 2008.

\bibitem{alldieck2019learning}
Thiemo Alldieck, Marcus Magnor, Bharat~Lal Bhatnagar, Christian Theobalt, and
  Gerard Pons-Moll.
\newblock Learning to reconstruct people in clothing from a single {RGB}
  camera.
\newblock In {\em {IEEE}/{CVF} Conference on Computer Vision and Pattern
  Recognition ({CVPR})}, 2019.

\bibitem{alldieck2018detailed}
Thiemo Alldieck, Marcus Magnor, Weipeng Xu, Christian Theobalt, and Gerard
  Pons-Moll.
\newblock Detailed human avatars from monocular video.
\newblock In {\em International Conference on 3D Vision (3DV)}, sep 2018.

\bibitem{alldieck2018video}
Thiemo Alldieck, Marcus Magnor, Weipeng Xu, Christian Theobalt, and Gerard
  Pons-Moll.
\newblock Video based reconstruction of {3D} people models.
\newblock In {\em {IEEE} Conf. on Computer Vision and Pattern Recognition},
  2018.

\bibitem{Anguelov2005scape}
Dragomir Anguelov, Praveen Srinivasan, Daphne Koller, Sebastian Thrun, Jim
  Rodgers, and James Davis.
\newblock Scape: Shape completion and animation of people.
\newblock {\em ACM Transactions on Graphics}, 24(3):408--416, July 2005.

\bibitem{Baran2007skin}
Ilya Baran and Jovan Popovi\'{c}.
\newblock Automatic rigging and animation of 3d characters.
\newblock In {\em SIGGRAPH}, SIGGRAPH '07, New York, NY, USA, 2007. ACM.

\bibitem{BogoBL015Detailed}
Federica Bogo, Michael~J. Black, Matthew Loper, and Javier Romero.
\newblock Detailed full-body reconstructions of moving people from monocular
  {RGB-D} sequences.
\newblock In {\em IEEE ICCV}, 2015.

\bibitem{Bogo2016keepsmpl}
Federica Bogo, Angjoo Kanazawa, Christoph Lassner, Peter Gehler, Javier Romero,
  and Michael~J. Black.
\newblock Keep it {SMPL}: Automatic estimation of {3D} human pose and shape
  from a single image.
\newblock In {\em IEEE ECCV}, Lecture Notes in Computer Science. Springer
  International Publishing, 2016.

\bibitem{Bradley:2008}
Derek Bradley, Tiberiu Popa, Alla Sheffer, Wolfgang Heidrich, and Tamy
  Boubekeur.
\newblock Markerless garment capture.
\newblock {\em ACM Trans. Graphics (Proc. SIGGRAPH)}, 27(3):99, 2008.

\bibitem{chen2015garment}
Xiaowu Chen, Bin Zhou, Feixiang Lu, Lin Wang, Lang Bi, and Ping Tan.
\newblock Garment modeling with a depth camera.
\newblock {\em ACM Transactions on Graphics (TOG)}, 34(6):203, 2015.

\bibitem{chen2016realtime}
Yin Chen, Zhi-Quan Cheng, Chao Lai, Ralph~R Martin, and Gang Dang.
\newblock Realtime reconstruction of an animating human body from a single
  depth camera.
\newblock {\em IEEE Transactions on Visualization and Computer Graphics},
  22(8):2000--2011, 2016.

\bibitem{Collet2015FVV}
Alvaro Collet, Ming Chuang, Pat Sweeney, Don Gillett, Dennis Evseev, David
  Calabrese, Hugues Hoppe, Adam Kirk, and Steve Sullivan.
\newblock High-quality streamable free-viewpoint video.
\newblock {\em ACM Trans. Graph.}, 34(4):69:1--69:13, July 2015.

\bibitem{danvevrek2017deepgarment}
R Dan{\v{e}}{\v{r}}ek, Endri Dibra, Cengiz {\"O}ztireli, Remo Ziegler, and
  Markus Gross.
\newblock Deepgarment: 3d garment shape estimation from a single image.
\newblock In {\em Computer Graphics Forum}, volume~36, pages 269--280. Wiley
  Online Library, 2017.

\bibitem{de2010stable}
Edilson De~Aguiar, Leonid Sigal, Adrien Treuille, and Jessica~K Hodgins.
\newblock Stable spaces for real-time clothing.
\newblock In {\em ACM Transactions on Graphics (TOG)}, volume~29, page 106.
  ACM, 2010.

\bibitem{dou2017motion2fusion}
Mingsong Dou, Philip Davidson, Sean Fanello, Sameh Khamis, Adarsh Kowdle,
  Christoph Rhemann, Vladimir Tankovich, and Shahram Izadi.
\newblock Motion2fusion: real-time volumetric performance capture.
\newblock 36:1--16, 11 2017.

\bibitem{dou2016fusion4d}
Mingsong Dou, Sameh Khamis, Yury Degtyarev, Philip Davidson, Sean~Ryan Fanello,
  Adarsh Kowdle, Sergio~Orts Escolano, Christoph Rhemann, David Kim, Jonathan
  Taylor, et~al.
\newblock Fusion4d: real-time performance capture of challenging scenes.
\newblock {\em ACM Transactions on Graphics}, 35(4):114, 2016.

\bibitem{gall2009motion}
Juergen Gall, Carsten Stoll, Edilson De~Aguiar, Christian Theobalt, Bodo
  Rosenhahn, and Hans-Peter Seidel.
\newblock Motion capture using joint skeleton tracking and surface estimation.
\newblock In {\em IEEE CVPR}, 2009.

\bibitem{gillette2015real}
Russell Gillette, Craig Peters, Nicholas Vining, Essex Edwards, and Alla
  Sheffer.
\newblock Real-time dynamic wrinkling of coarse animated cloth.
\newblock In {\em Proceedings of the 14th ACM SIGGRAPH/Eurographics Symposium
  on Computer Animation}, pages 17--26. ACM, 2015.

\bibitem{goldenthal2007efficient}
Rony Goldenthal, David Harmon, Raanan Fattal, Michel Bercovier, and Eitan
  Grinspun.
\newblock Efficient simulation of inextensible cloth.
\newblock In {\em ACM Transactions on Graphics (TOG)}, volume~26, page~49. ACM,
  2007.

\bibitem{guan2012drape}
Peng Guan, Loretta Reiss, David~A Hirshberg, Alexander Weiss, and Michael~J
  Black.
\newblock Drape: Dressing any person.
\newblock {\em ACM Trans. Graph.}, 31(4):35--1, 2012.

\bibitem{guo2015robust}
Kaiwen Guo, Feng Xu, Yangang Wang, Yebin Liu, and Qionghai Dai.
\newblock Robust non-rigid motion tracking and surface reconstruction using l0
  regularization.
\newblock In {\em IEEE ICCV}, 2015.

\bibitem{guo2017real}
Kaiwen Guo, Feng Xu, Tao Yu, Xiaoyang Liu, Qionghai Dai, and Yebin Liu.
\newblock Real-time geometry, albedo and motion reconstruction using a single
  rgbd camera.
\newblock {\em ACM Transactions on Graphics}, 2017.

\bibitem{habermann2019TOG}
Marc Habermann, Weipeng Xu, , Michael Zollhoefer, Gerard Pons-Moll, and
  Christian Theobalt.
\newblock Livecap: Real-time human performance capture from monocular video.
\newblock {\em Transactions on Graphics (ToG) 2019}, oct 2019.

\bibitem{tung2017self}
Yu Hsiao Fish~Tung, Wei~Tung Hsiao, Yumer Ersin, and Fragkiadaki Katerina.
\newblock Self-supervised learning of motion capture.
\newblock In {\em Neural Information Processing Systems (NIPS)}, 2017.

\bibitem{innmann2016volume}
Matthias Innmann, Michael Zollh{\"o}fer, Matthias Nie{\ss}ner, Christian
  Theobalt, and Marc Stamminger.
\newblock Volumedeform: Real-time volumetric non-rigid reconstruction.
\newblock In {\em IEEE ECCV}, 2016.

\bibitem{Jakob2015Instant}
Wenzel Jakob, Marco Tarini, Daniele Panozzo, and Olga Sorkine-Hornung.
\newblock Instant field-aligned meshes.
\newblock {\em ACM Transactions on Graphics (Proceedings of SIGGRAPH ASIA)},
  34(6), Nov. 2015.

\bibitem{jiang2017anisotropic}
Chenfanfu Jiang, Theodore Gast, and Joseph Teran.
\newblock Anisotropic elastoplasticity for cloth, knit and hair frictional
  contact.
\newblock {\em ACM Transactions on Graphics (TOG)}, 36(4):152, 2017.

\bibitem{hmrKanazawa17}
Angjoo Kanazawa, Michael~J. Black, David~W. Jacobs, and Jitendra Malik.
\newblock End-to-end recovery of human shape and pose.
\newblock In {\em Computer Vision and Pattern Regognition (CVPR)}, 2018.

\bibitem{kavan2011physics}
Ladislav Kavan, Dan Gerszewski, Adam~W Bargteil, and Peter-Pike Sloan.
\newblock Physics-inspired upsampling for cloth simulation in games.
\newblock In {\em ACM Transactions on Graphics (TOG)}, volume~30, page~93. ACM,
  2011.

\bibitem{Kazhdan:2006:PSR:1281957.1281965}
Michael Kazhdan, Matthew Bolitho, and Hugues Hoppe.
\newblock Poisson surface reconstruction.
\newblock In {\em Proceedings of the Fourth Eurographics Symposium on Geometry
  Processing}, SGP '06, pages 61--70, Aire-la-Ville, Switzerland, Switzerland,
  2006. Eurographics Association.

\bibitem{kim2013near}
Doyub Kim, Woojong Koh, Rahul Narain, Kayvon Fatahalian, Adrien Treuille, and
  James~F O'Brien.
\newblock Near-exhaustive precomputation of secondary cloth effects.
\newblock {\em ACM Transactions on Graphics (TOG)}, 32(4):87, 2013.

\bibitem{Meekyoung:siggraph}
Meekyoung Kim, Gerard Pons-Moll, Sergi Pujades, Sungbae Bang, Jinwwok Kim,
  Michael Black, and Sung-Hee Lee.
\newblock Data-driven physics for human soft tissue animation.
\newblock {\em ACM Transactions on Graphics, (Proc. SIGGRAPH)}, 36(4), 2017.

\bibitem{Lahner_2018_ECCV}
Zorah Lahner, Daniel Cremers, and Tony Tung.
\newblock Deepwrinkles: Accurate and realistic clothing modeling.
\newblock In {\em The European Conference on Computer Vision (ECCV)}, September
  2018.

\bibitem{chao2018ArticulatedFusion}
Chao Li, Zheheng Zhang, and Xiaohu Guo.
\newblock Articulatedfusion: Real-time reconstruction of motion, geometry and
  segmentation using a single depth camera.
\newblock In {\em ECCV}, 2018.

\bibitem{li2009robust}
Hao Li, Bart Adams, Leonidas~J Guibas, and Mark Pauly.
\newblock Robust single-view geometry and motion reconstruction.
\newblock In {\em ACM Transactions on Graphics}, volume~28, page 175. ACM,
  2009.

\bibitem{FLAME:SiggraphAsia2017}
Tianye Li, Timo Bolkart, Michael.~J. Black, Hao Li, and Javier Romero.
\newblock Learning a model of facial shape and expression from {4D} scans.
\newblock {\em ACM Transactions on Graphics, (Proc. SIGGRAPH Asia)}, 36(6),
  2017.

\bibitem{LIP2018}
Xiaodan Liang, Ke Gong, Xiaohui Shen, and Liang Lin.
\newblock Look into person: Joint body parsing amp; pose estimation network and
  a new benchmark.
\newblock {\em IEEE Transactions on Pattern Analysis and Machine Intelligence},
  pages 1--1, 2018.

\bibitem{liu2011markerless}
Yebin Liu, Carsten Stoll, Juergen Gall, Hans-Peter Seidel, and Christian
  Theobalt.
\newblock Markerless motion capture of interacting characters using multi-view
  image segmentation.
\newblock In {\em IEEE CVPR}, 2011.

\bibitem{Loper2015SMPL}
Matthew Loper, Naureen Mahmood, Javier Romero, Gerard Pons-Moll, and Michael~J.
  Black.
\newblock {SMPL}: A skinned multi-person linear model.
\newblock {\em ACM Transactions on Graphics (Proc. SIGGRAPH Asia)},
  34(6):248:1--248:16, Oct. 2015.

\bibitem{muller2008hpbd}
Matthias M{\"u}ller.
\newblock Hierarchical position based dynamics.
\newblock 2008.

\bibitem{muller2007PBD}
Matthias M{\"u}ller, Bruno Heidelberger, Marcus Hennix, and John Ratcliff.
\newblock Position based dynamics.
\newblock {\em Journal of Visual Communication and Image Representation},
  18(2):109--118, 2007.

\bibitem{neophytou2014layered}
Alexandros Neophytou and Adrian Hilton.
\newblock A layered model of human body and garment deformation.
\newblock In {\em 3D Vision (3DV), 2014 2nd International Conference on},
  volume~1, pages 171--178. IEEE, 2014.

\bibitem{newcombe2015dynamic}
Richard~A. Newcombe, Dieter Fox, and Steven~M. Seitz.
\newblock Dynamicfusion: Reconstruction and tracking of non-rigid scenes in
  real-time.
\newblock In {\em IEEE CVPR}, 2015.

\bibitem{omran2018neural}
Mohamed Omran, Christop Lassner, Gerard Pons-Moll, Peter Gehler, and Bernt
  Schiele.
\newblock Neural body fitting: Unifying deep learning and model based human
  pose and shape estimation.
\newblock In {\em International Conf. on 3D Vision}, 2018.

\bibitem{pavlakos2018humanshape}
Georgios Pavlakos, Luyang Zhu, Xiaowei Zhou, and Kostas Daniilidis.
\newblock Learning to estimate 3{D} human pose and shape from a single color
  image.
\newblock In {\em {IEEE} Conf. on Computer Vision and Pattern Recognition},
  2018.

\bibitem{Pons-Moll:Siggraph2017}
Gerard Pons-Moll, Sergi Pujades, Sonny Hu, and Michael Black.
\newblock {ClothCap}: Seamless {4D} clothing capture and retargeting.
\newblock {\em ACM Transactions on Graphics, (Proc. SIGGRAPH)}, 36(4), 2017.

\bibitem{PonsMoll2015Dyna}
Gerard Pons-Moll, Javier Romero, Naureen Mahmood, and Michael~J. Black.
\newblock Dyna: A model of dynamic human shape in motion.
\newblock {\em ACM Transactions on Graphics, (Proc. SIGGRAPH)},
  34(4):120:1--120:14, Aug. 2015.

\bibitem{Pons-Moll:IJCV:2015}
Gerard Pons-Moll, Jonathan Taylor, Jamie Shotton, Aaron Hertzmann, and Andrew
  Fitzgibbon.
\newblock Metric regression forests for correspondence estimation.
\newblock {\em International Journal of Computer Vision}, pages 1--13, 2015.

\bibitem{Popa:2009}
Tiberiu Popa, Qingnan Zhou, Derek Bradley, Vladislav Kraevoy, Hongbo Fu, Alla
  Sheffer, and Wolfgang Heidrich.
\newblock Wrinkling captured garments using space-time data-driven deformation.
\newblock {\em Computer Graphics Forum (Proc. Eurographics)}, 28(2):427--435,
  2009.

\bibitem{provot97collision}
Xavier Provot.
\newblock Collision and self-collision handling in cloth model dedicated to
  design garments.
\newblock {\em Graphics Interface. 97}, 1997.

\bibitem{provot1995deformation}
Xavier Provot et~al.
\newblock Deformation constraints in a mass-spring model to describe rigid
  cloth behaviour.
\newblock In {\em Graphics interface}, pages 147--147. Canadian Information
  Processing Society, 1995.

\bibitem{MANO:SIGGRAPHASIA:2017}
Javier Romero, Dimitrios Tzionas, and Michael~J. Black.
\newblock Embodied hands: Modeling and capturing hands and bodies together.
\newblock {\em ACM Transactions on Graphics, (Proc. SIGGRAPH Asia)}, 36(6),
  Nov. 2017.

\bibitem{rosenhahn2007system}
Bodo Rosenhahn, Uwe Kersting, Katie Powell, Reinhard Klette, Gisela Klette, and
  Hans-Peter Seidel.
\newblock A system for articulated tracking incorporating a clothing model.
\newblock {\em Machine Vision and Applications}, 18(1):25--40, 2007.

\bibitem{selle2009robust}
Andrew Selle, Jonathan Su, Geoffrey Irving, and Ronald Fedkiw.
\newblock Robust high-resolution cloth using parallelism, history-based
  collisions, and accurate friction.
\newblock {\em IEEE Transactions on Visualization and Computer Graphics},
  15(2):339--350, 2009.

\bibitem{slavcheva2017cvpr}
Miroslava Slavcheva, Maximilian Baust, Daniel Cremers, and Slobodan Ilic.
\newblock {KillingFusion: Non-rigid 3D Reconstruction without Correspondences}.
\newblock In {\em IEEE CVPR}, 2017.

\bibitem{slavcheva2018cvpr}
Miroslava Slavcheva, Maximilian Baust, and Slobodan Ilic.
\newblock {SobolevFusion: 3D Reconstruction of Scenes Undergoing Free Non-rigid
  Motion}.
\newblock In {\em IEEE/CVF Conference on Computer Vision and Pattern
  Recognition (CVPR)}, 2018.

\bibitem{stoll2010video}
Carsten Stoll, Juergen Gall, Edilson De~Aguiar, Sebastian Thrun, and Christian
  Theobalt.
\newblock Video-based reconstruction of animatable human characters.
\newblock In {\em ACM Transactions on Graphics (TOG)}, volume~29, page 139.
  ACM, 2010.

\bibitem{Sumner2007embededed}
Robert~W. Sumner, Johannes Schmid, and Mark Pauly.
\newblock Embedded deformation for shape manipulation.
\newblock SIGGRAPH '07, New York, NY, USA, 2007. ACM.

\bibitem{taylor2012vitruvian}
Jonathan Taylor, Jamie Shotton, Toby Sharp, and Andrew Fitzgibbon.
\newblock The vitruvian manifold: Inferring dense correspondences for one-shot
  human pose estimation.
\newblock In {\em IEEE CVPR}, 2012.

\bibitem{terzopoulos1987elastically}
Demetri Terzopoulos, John Platt, Alan Barr, and Kurt Fleischer.
\newblock Elastically deformable models.
\newblock {\em ACM Siggraph Computer Graphics}, 21(4):205--214, 1987.

\bibitem{vlasic2008articulated}
Daniel Vlasic, Ilya Baran, Wojciech Matusik, and Jovan Popovi{\'c}.
\newblock Articulated mesh animation from multi-view silhouettes.
\newblock In {\em ACM Transactions on Graphics}, volume~27, page~97. ACM, 2008.

\bibitem{wang2010example}
Huamin Wang, Florian Hecht, Ravi Ramamoorthi, and James~F O'Brien.
\newblock Example-based wrinkle synthesis for clothing animation.
\newblock In {\em Acm Transactions on Graphics (TOG)}, volume~29, page 107.
  ACM, 2010.

\bibitem{wuhrer2014estimation}
Stefanie Wuhrer, Leonid Pishchulin, Alan Brunton, Chang Shu, and Jochen Lang.
\newblock Estimation of human body shape and posture under clothing.
\newblock {\em Computer Vision and Image Understanding}, 127:31--42, 2014.

\bibitem{MonoPerfCap_SIGGRAPH2018}
Weipeng Xu, Avishek Chatterjee, Michael Zollhoefer, Helge Rhodin, Dushyant
  Mehta, Hans-Peter Seidel, and Christian Theobalt.
\newblock Monoperfcap: Human performance capture from monocular video, 2018.

\bibitem{xu2014sensitivity}
Weiwei Xu, Nobuyuki Umetani, Qianwen Chao, Jie Mao, Xiaogang Jin, and Xin Tong.
\newblock Sensitivity-optimized rigging for example-based real-time clothing
  synthesis.
\newblock {\em ACM Trans. Graph.}, 33(4):107--1, 2014.

\bibitem{yang2016estimation}
Jinlong Yang, Jean-S{\'e}bastien Franco, Franck H{\'e}troy-Wheeler, and
  Stefanie Wuhrer.
\newblock {Estimation of Human Body Shape in Motion with Wide Clothing}.
\newblock In {\em European Conf. on Computer Vision}, Amsterdam, Netherlands,
  2016.

\bibitem{yang2018analyzing}
Jinlong Yang, Jean-S{\'e}bastien Franco, Franck H{\'e}troy-Wheeler, and
  Stefanie Wuhrer.
\newblock Analyzing clothing layer deformation statistics of 3d human motions.
\newblock In Vittorio Ferrari, Martial Hebert, Cristian Sminchisescu, and Yair
  Weiss, editors, {\em Computer Vision -- ECCV 2018}, pages 245--261, Cham,
  2018. Springer International Publishing.

\bibitem{ye2012performance}
Genzhi Ye, Yebin Liu, Nils Hasler, Xiangyang Ji, Qionghai Dai, and Christian
  Theobalt.
\newblock Performance capture of interacting characters with handheld kinects.
\newblock In {\em ECCV}. 2012.

\bibitem{ye2014real}
Mao Ye and Ruigang Yang.
\newblock Real-time simultaneous pose and shape estimation for articulated
  objects using a single depth camera.
\newblock In {\em IEEE CVPR}, 2014.

\bibitem{tao2017BodyFusion}
Tao Yu, Kaiwen Guo, Feng Xu, Yuan Dong, Zhaoqi Su, Jianhui Zhao, Jianguo Li,
  Qionghai Dai, and Yebin Liu.
\newblock Bodyfusion: Real-time capture of human motion and surface geometry
  using a single depth camera.
\newblock In {\em IEEE ICCV}, 2017.

\bibitem{tao2018DoubleFusion}
Tao Yu, Zerong Zheng, Kaiwen Guo, Jianhui Zhao, Qionghai Dai, Hao Li, Gerard
  Pons-Moll, and Yebin Liu.
\newblock Doublefusion: Real-time capture of human performances with inner body
  shapes from a single depth sensor.
\newblock In {\em The IEEE International Conference on Computer Vision and
  Pattern Recognition(CVPR)}. IEEE, June 2018.

\bibitem{zhang2017detailed}
Chao Zhang, Sergi Pujades, Michael Black, and Gerard Pons-Moll.
\newblock Detailed, accurate, human shape estimation from clothed {3D} scan
  sequences.
\newblock In {\em IEEE CVPR}, 2017.

\bibitem{zhou2013garment}
Bin Zhou, Xiaowu Chen, Qiang Fu, Kan Guo, and Ping Tan.
\newblock Garment modeling from a single image.
\newblock In {\em Computer graphics forum}, volume~32, pages 85--91. Wiley
  Online Library, 2013.

\end{thebibliography}
